\documentclass[a4paper]{IEEEtran}
\usepackage{cite}
\usepackage{amsmath,amssymb}
\usepackage{algorithmic}
\usepackage{graphicx}
\usepackage{times}
\usepackage{url}
\usepackage{bm}
\usepackage{multirow}
\usepackage{caption}
\usepackage{subcaption}
\usepackage{listings}
\usepackage[utf8]{inputenc}
\usepackage[vietnamese,english]{babel}
\usepackage{CJKutf8}
\usepackage[utf8]{inputenc}
\usepackage[T1]{fontenc}
\usepackage{color}

\definecolor{codegreen}{rgb}{0,0.6,0}
\definecolor{codegray}{rgb}{0.5,0.5,0.5}
\definecolor{codepurple}{rgb}{0.58,0,0.82}
\definecolor{backcolour}{rgb}{0.95,0.95,0.92}
 
\lstdefinestyle{mystyle}{
    backgroundcolor=\color{backcolour},   
    commentstyle=\color{codegreen},
    keywordstyle=\color{magenta},
    numberstyle=\tiny\color{codegray},
    stringstyle=\color{codepurple},
    basicstyle=\footnotesize,
    breakatwhitespace=false,         
    breaklines=true,                 
    captionpos=b,                    
    keepspaces=true,                                  
    showspaces=false,                
    showstringspaces=false,
    showtabs=false,                  
    tabsize=2
}
 
\newcommand{\JP}[1]{\begin{CJK}{UTF8}{min}#1\end{CJK}}
\newcommand{\dotproduct}{\raisebox{.3ex}{\tiny$\bullet$}} 

\def\BibTeX{{\rm B\kern-.05em{\sc i\kern-.025em b}\kern-.08em
    T\kern-.1667em\lower.7ex\hbox{E}\kern-.125emX}}
\begin{document}

\title{Combining Advanced Methods in Japanese-Vietnamese Neural Machine Translation}

\author{
\begin{tabular}{c} \IEEEauthorblockN{Thi-Vinh Ngo} \\ \IEEEauthorblockA{
\textit{University of Information and Communication Technology}\\
\textit{TNU, Vietnam}\\ 
ntvinh@ictu.edu.vn\\}\\
\IEEEauthorblockN{Phuong-Thai Nguyen} \\ \IEEEauthorblockA{
\textit{University of Engineering and Technology}\\
\textit{VNU, Vietnam}\\
npthai@vnu.edu} \end{tabular} \and
\begin{tabular}{c} \IEEEauthorblockN{Thanh-Le Ha} \\ \IEEEauthorblockA{
\textit{Institute of Anthropomatics and Robotics}\\
\textit{KIT, Germany}\\
thanh-le.ha@kit.edu\\}\\
 \IEEEauthorblockN{Le-Minh Nguyen} \\ \IEEEauthorblockA{
 \textit{School of Information Science}\\
  \textit{JAIST, Japan} \\
nguyenml@jaist.ac.jp}\end{tabular}\\
}

\maketitle

\begin{abstract}
Neural machine translation (NMT) systems have recently obtained state-of-the art in many machine translation systems between popular language pairs because of the availability of data. For low-resourced language pairs, there are few researches in this field due to the lack of bilingual data. In this paper, we attempt to build the first NMT systems  for a low-resourced language pairs:Japanese-Vietnamese. We have also shown significant improvements when combining advanced methods to reduce the adverse impacts of data sparsity and improve the quality of NMT systems. In addition, we proposed a variant of Byte-Pair Encoding algorithm to perform effective word segmentation for Vietnamese texts and alleviate the rare-word problem that persists in NMT systems.
\end{abstract}

\section{Introduction}
\label{introduction}
Neural machine translation (NMT)\cite{Bahdanau2014,Sutskever2014} is widely applied for machine translation (MT) in recent years and focuses on popular language pairs such as  English$\leftrightarrow$French, English$\leftrightarrow$German, English$\leftrightarrow$Chinese or English$\leftrightarrow$Japanese. NMT has obtained state-of-the-art performance on those language pairs compared to the traditional statistical machine translation (SMT) when given enough data\cite{wu2016google,cettolo2016iwslt}. Furthermore, due to the ability of feature learning, NMT systems can be trained end-to-end with pure parallel texts and minimal linguistic knowledge of the languages involved. Thus it makes training NMT for a new language pair much easier, more scalable and robust. Nevertheless, NMT has not been employed in many low-resourced language pairs since in those scenarios, data scarcity often limits the learning ability of neural methods. In contrast, combinating complicated linguistic-driven features in a typical log-linear framework still keeps SMT the best approach in many translation directions but also hard to apply to new domains or to other language pairs. 

In this paper we attempt to build NMT systems for such a low-resourced language pair: Japanese$\leftrightarrow$Vietnamese. Our aim is to set the first and reasonable NMT systems that can be reproducible in order to serve as the baseline for further researches in the direction\footnote{All the corpora and codes used in the experiments will be published later.}. Furthermore, we conduct experiments using some advanced methods to improve the quality of the systems. An important criteria for those methods is that they must be scalable and language-independent as much as possible. The criteria ensures the basic principle of NMT as well as the reproducibility of the systems. On the other hand, the methods are chosen in the direction that they would help alleviate the data sparsity problem of NMT when being applied in this low-resourced setting. 

Specifically, to deal with rare-word translation problems, we experiment with translation units in different levels: subword, word and beyond. In morphological-rich languages such as English or German, using subword as the translation units is often suitable since neural methods are able to induce meaning or linguistic function of sub-components constituting a word. Byte-Pair Encoding (BPE)\cite{Sennrich2016a} is a simple unsupervised technique to do subword segmentation and it has great effects when applied to NMT training. Japanese and Vietnamese (and some other Asian languages), however, have different word segmentation issues. Hence, it would be difficult to apply BPE directly to the texts in those languages and build NMT systems for subword translation without any modification. In this paper, we experiment different segmentation methods for both languages and also propose a variant of BPE algorithm to learn translation units for Vietnamese in an unsupervised way. 
  
We also attempt to increase the amount of training data by using back-translated texts or mix-source data just from our small available corpus. Those data augmentation approaches have shown their effectiveness on various NMT systems, especially in under-resourced scenarios. While back translation technique is used to generate synthetic data from monolingual corpora, mix-source technique utilizes human-quality corpora in a multilingual setting, leveraging the transfer learning ability across languages. Both are simple but elegantly model the relevant noise needed in training neural architectures in such low-resourced situations.

The main contributions of this paper are:
\begin{itemize}
\item We created the first NMT systems for Japanese$\leftrightarrow$Vietnamese and released the dataset as well as the codes to reproduce the experiments. 
\item We conducted several ways of segmentation and proposed a variant of BPE algorithm for Vietnamese, which does not need any labeled data or linguistic resource.
\item We applied elegant data augmentation methods in order to reduce the severeness of data sparsity problem in training NMT systems using a small dataset of Japanese-Vietnamese.
\end{itemize}
 
\section{Neural Machine Translation} 
\label{nmt}
In this section, we will describe the general architecture of NMT as a kind of sequence-to-sequence modeling framework. In this kind of sequence-to-sequence modeling framework, often there is an encoder trying to encode context information from the input sequence and a decoder to generate one item of the output sequence at a time based on the context of both input and output sequences. Besides, an additional component, named attention, exists in between, deciding which parts of the input sequence the decoder should  pay attention in order to choose which to output next. In other words, this attention component calculates the context relevant to the decision of the decoder at the considering time.  Those components as a whole constitute a large trainable neural architecture called the famous attention-based encoder-decoder framework. This framework becomes popular in many sequence-to-sequence tasks.

In the field of machine translation, using the attention-based encoder-decoder framework is referred to as Neural Machine Translation approach. First presented in \cite{Bahdanau2014}, the encoder and decoder in NMT are recurrent-based, which each hidden unit in those components is a recurrent unit like Long Short-term Memory (LSTM)~\cite{HochreiterLSTM} or Gated Recurrent Unit(GRU) \cite{Cho2014}. Later, the encoder or decoder can also be a convolutional architecture, as in \cite{gehring2017convolutional}. Recently, \cite{vaswani2017attention} introduces transformer architecture, in which both the encoder and decoder are a special variant of attention mechanism, called self attention. In this paper, we will briefly explain the Recurrent NMT as we utilize it in our experiments.

The Recurrent NMT model follows the attention-based architecture proposed by \cite{Bahdanau2014}. The bidirectional recurrent encoder reads every words $x_{i}$ of a source sentence $\bm{x}=\{x_1,...,x_n\}$ and encodes a representation $\bm{s}$ of the sentence into a fixed-length vector $\bm{h}_i$ 
concatenated from those of the forward and backward directions:

\[
\begin{aligned}
& \bm{h}_i=[\overrightarrow{\bm{h}}_i,\overleftarrow{\bm{h}}_i] \\
& \overrightarrow{\bm{h}}_i=f(\overrightarrow{\bm{h}}_{i-1},\bm{s}) \\
& \overleftarrow{\bm{h}}_i=f(\overleftarrow{\bm{h}}_{i+1},\bm{s}) \\
& \bm{s}=\bm{E}_s~\dotproduct~\bm{x}_i \\
\end{aligned}
\]

Here $\bm{x}_i$ is the one-hot vector of the word $x_i$ and $\bm{E}_s$ is the word embedding matrix which is shared across the source words. $f$ is the recurrent unit computing the current hidden state of the encoder based on the previous hidden state. $\bm{h}_i$ is then called an \textit{annotation vector}, 
which represent the information of the source sentence up to the time $i$ from both forward and backward directions. 

Those annotation vectors of the source sentences are combined in the attention layers in a way that the resulted vector encodes the source context relevant to the considering target word the decoder should produce. Intuitively, a relevance between the previous target word $\bm{z}_{j-1}$ and the annotation vectors  $\bm{h}_i$ corresponding to the source words $x_i$ can be used to form some attention scenario:
\[
\begin{aligned} \label{eq:1}
& rel(\bm{z}_{j-1},\bm{h}_i) = \bm{v}_a~\dotproduct~\tanh(\bm{W}_a~\dotproduct~\bm{z}_{j-1} + \bm{U}_a~\dotproduct~\bm{h}_i) \\
& \alpha_{ij} = \displaystyle  \frac{\exp(rel(\bm{z}_{j-1},\bm{h}_i))}{\sum_{i'} \exp(rel(\bm{z}_{j-1},\bm{h}_{i'}))}~~\text{and}~~\bm{c}_j = \displaystyle \sum_{i}{\alpha_{ij}\bm{h}_i}\\
\end{aligned}
\]

This specific attention mechanism, originally called \textit{alignment model} in ~\cite{Bahdanau2014},
has been employed as a simple feedforward network with the first layer is a learnable layer via adaptation factors $\bm{v}_a$,$\bm{W}_a$ and $\bm{U}_a$.
The relevance scores $rel$ are then normalized into attention weights $\alpha_{ij}$ 
and the context vector $\bm{c}_j$ is calculated as the weighted sum of all annotation vectors $\bm{h}_i$. 
Depending on how much attention the target word at time $j$ put on the source states $\bm{h}_i$, a soft alignment is learned and a source context $\bm{c}_j$ at time $j$ is calculated prior to the prediction of the decoder. 


Similar to the encoder, the recurrent decoder recursively generates one target word $y_{j}$ to form a translated target sentence $\bm{y}=\{y_1,...,y_m\}$ in the end. At time $j$, it takes the previous hidden state of the decoder $\bm{z}_{j-1}$, the previous embedded word representation $\bm{t}_{j-1}$ 
and the time-specific context vector $\bm{c}_j$ as inputs to calculate the current hidden state $\bm{z}_{j}$:

\[
\begin{aligned}
& \bm{z}_{j}=g(\bm{z}_{j-1}, \bm{t}_{j-1}, \bm{c}_j) \\
& \bm{t}_{j-1} = \bm{E}_t~\dotproduct~\bm{y}_{j-1}
\end{aligned}
\]

Again, $g$ is the recurrent activation function of the decoder and $\bm{E}_t$ is the shared word embedding matrix of the target sentences. 

Given the parallel corpus consisting of $N$ training examples $\{(\bm{x}^{(n)},\bm{y}^{(n)})\}_{n=1}^N$, the objective is to maximize the conditional log-probability of the correct translation given a source sentence with respect to the parameters $\theta$ of the whole model:

$$ L(\theta)= \displaystyle \arg\max_{\theta} \frac{1}{N} \sum_{n=1}^{N}\log p(\bm{y}^{(n)}|\bm{x}^{(n)},\theta)$$



\section{Subword Translation}
\label{subword}
One of the most severe problems of NMT is dealing with the rare words, 
which are not in the short lists of the vocabularies, i.e. out-of-vocabulary (OOV) words, or do not appear in the training set at all. On one hand, we would like to have fewer OOV words by increasing the size of the short lists. On the other hand, we need our neural network to learn fast and has a good generalization capability on the unseen words as well.

As explained shortly in the introduction, for many languages, using subword instead of word as a \textbf{\textit{translation unit}} (\textbf{\textit{TU}}) has been shown that it is not only effective on reducing vocabulary sizes, thus alleviating the computing burden on the large soft-max layer as well as reducing a substantial number of parameters to be learnt, but also has the ability to generate unseen words. In those languages, a word can be a compound word or comprised by sub-components, each has its own raw meaning or contains morphological information. Segmenting words into sub-components allows NMT to learn to translate them with considerably fewer data. For example, it is definitely less chance to see this \textbf{popular} German word ``\textit{Wohnungsreinigung}" (English equivalence: ``\textit{house cleaning}'') than its sub-components ``\textit{Wohnung}'' (i.e. ``\textit{house}'' or ``\textit{flat}'') and ``\textit{reinigung}'' (i.e. ``\textit{cleaning}'') in a middle-sized German-English parallel corpus. Instead, NMT can observe and translate those sub-components (``\textit{Wohnung}'' and ``\textit{reinigung}'') and combine their translations to generate the unseen words (``\textit{house cleaning}''). This is achieved by segmenting words into subword units using segmentation techniques in the preprocessing phase prior to translation. There are several segmentation methods; Some are the complicate ones which require linguistic resources or human-crafted rules. Thus, they are not language-independent and expensive to obtain for low-resourced languages.

Byte-Pair Encoding, otherwise, is a simple but robust technique to do subword segmentation. Since it is an unsupervised and fast technique, it has great effects when applied to build NMT systems for morphologically rich languages. BPE is originally proposed in \cite{gage1994} as a data compression technique by iteratively replaces the most frequent pair of bytes in a sequence with a single, unused byte.~\cite{Sennrich2016a} realized Gage's algorithm for word segmentation by merging frequent characters \textit{inside a word} instead of merging frequent pairs of bytes in a whole file (sequence of bytes). Being applied in translation from and to morphological rich languages, it can automatically induce sub-components of a word (i.e. sequence of characters) which bear some meaning or morphological function without knowing much about linguistic characteristics of those languages. Hence, the \textbf{\textit{TU}} of the NMT used in those languages is at a smaller level of word, i.e. subword.    

On the other hand, Japanese and Vietnamese are different to those languages in the way of how we consider \textit{a word}. In Vietnamese, the \textbf{\textit{TU}}, normally considered as \textit{a word}, which often separated to each others by white spaces, is not a word in linguistic term since it does not really have its own meaning. Therefore, applying BPE to segment those \textbf{\textit{TU}}s into smaller units without any modification is not suitable for Vietnamese. In case of Japanese, there are no space to separate written texts into \textbf{\textit{TU}}s. Thus, for Vietnamese and Japanese as well as for the languages having similar problem, before applying BPE, it is necessary to have some preprocessing step to tokenize the texts into words. 

\subsection{Vietnamese Tokenization}
\foreignlanguage{Vietnamese}{
From the linguistic point of view, each sequence of characters between two white spaces in Vietnamese texts cannot be considered as a word since it does not always have a full meaning to stand alone. For example, in the sentence ``\textit{hôm nay là sinh nhật của tôi}'' (English equivalence: ``\textit{Today is my birthday}''), ``\textit{hôm}'' and ``\textit{nay}'' are not two words, they together form a word, which means ``\textit{today}''. Nevertheless, ``\textit{hôm}'' and ``\textit{nay}'' somehow still bear some meaning: ``\textit{hôm}''-``\textit{day}'', ``\textit{nay}''-``\textit{now}''. Similarly, ``\textit{sinh}''-``\textit{birth}'' and ``\textit{nhật}''-``\textit{date}'' also form the word ``\textit{sinh nhật}''-``\textit{birthday}'' but they are not two distinct words. We could also call them subwords.}

In many Vietnamese processing tasks such as Part-of-Speech Tagging, Syntax Parsing or Chunking, there requires a step to concatenate those subwords to make a word since in those tasks, word is necessarily considered as the smallest unit to be processed. This step is normally referred to as \textit{word segmentation}. There are various word segmentation methods; The best ones are using machine learning approaches to learn from a labeled corpus. It makes the tasks hard and expensive to be applied in other domains. Furthermore, the translation unit in Machine Translation does not need to be a word but can be a subword or sequence of subwords if it have its own meaning.

With this observation in mind, if we consider a subword, i.e. the sequence of characters between two white spaces, as a byte and a sentence as a sequence of bytes, we can apply the BPE algorithm straight-forward: We iteratively find and concatenate the most frequent pair of subwords ($w_1$,$w_2$) and replace it by an unseen subwords $w_1\_w_2$. We do not merge $w_1$ with $w_1$ if one of them are digits or punctuations or other special symbols. The BPE learning algorithm has an arguments which is the minimum value of frequency. In practice, we set the minimum frequency is 2. Listing~1 presents this variant of BPE, which we call \textbf{\textit{VNBPE}}.

\begin{table*}[t]
\begin{center}
\begin{tabular}{|c|l|l|l|}
\hline \hline
No. & Vietnamese phrases & Segmentation using pyvi’s algorithm & Segmentation using VN\_BPE  algorithm\\
\hline
1 & \foreignlanguage{Vietnamese}{sẽ kết thúc} & \foreignlanguage{Vietnamese}{sẽ kết\_thúc} & \foreignlanguage{Vietnamese}{sẽ\_kết\_thúc} \\
\hline
2 & \foreignlanguage{Vietnamese}{sự tập trung} & \foreignlanguage{Vietnamese}{sự tập\_trung} & \foreignlanguage{Vietnamese}{sự\_tập\_trung} \\
\hline
3 & \foreignlanguage{Vietnamese}{một đống} & \foreignlanguage{Vietnamese}{một đống} & \foreignlanguage{Vietnamese}{một\_đống} \\
\hline
4 & \foreignlanguage{Vietnamese}{vào lĩnh vực} & \foreignlanguage{Vietnamese}{vào lĩnh\_vực} & \foreignlanguage{Vietnamese}{vào\_lĩnh\_vực} \\
\hline 
5 & \foreignlanguage{Vietnamese}{bằng máy bay} & \foreignlanguage{Vietnamese}{bằng máy\_bay} & \foreignlanguage{Vietnamese}{bằng\_máy\_bay} \\
\hline \hline
\end{tabular}
\label{tab2}
\end{center}
\caption{\label{vbpe}Comparing a decent segmentation algorithm and VN\_BPE.}
\end{table*}

\noindent\begin{minipage}{\linewidth}
\begin{lstlisting}[language=Python, title=The proposed variant of BPE for Vietnamese.]
 def get_most_freq_pair(text,min_freq):
   dict = {}
   dumpWord = {set_of_separate_symbols}
   for line in text: 
     w1 = the_first_word_in_the_line  
     for w2 = each_word_in_the_line: 
       if w1,w2 not in dumpWord 
       and w1 is not a number: 
         dict[w1,w2] += 1 
         get_next_pair_in_the_line()
         w1 = w2      
   return (all pairs has freq > min_freq)
   
 def update_pairs(pair,text,codes_file): 
   original_word = pair[0] + " " + pair[1]
   replaced_word = pair[0] + "_" + pair[1]
   input_file.replace(rew,orw) 
   write_replaced_word_to_codes_file()
   return input_file 
   
 ### MAIN PART ###    
 min_freq=2
 open(input_file) to read
 open(codes_file) and (output_file) to write

 pairs = get_most_freq_pair(input_file,min_freq)
 arrange_pairs_for_decreasing_of_frequency()
 for each (freq,pair) in pairs: 
   update_pairs(pair,input_file,codes_file)
 output_file=input_file 
 close_all_files()
\end{lstlisting}
\end{minipage}
Compared to other word segmentation methods which require training on labeled data, our \textbf{\textit{VNBPE}} is a simple unsupervised method, alike to its original BPE algorithm. Table~\ref{vbpe} shows the outputs of one decent word segmentation method and our \textbf{\textit{VNBPE}}.

\subsection{Japanese Tokenization}
In a Japanese written text, there could be a mixture of three different types of scripts: Chinese characters (\textit{kanji}) and the other two syllabic scripts: \textit{hiragana} and \textit{katakana}. Each of \textit{kanji} characters can be loosely considered as a subword that we mentioned in the previous section. In the meanwhile, each of \textit{hiragana} or \textit{katakana} characters can be considered as a latin character in English or Vietnamese. In addition, there is no space in the Japanese written texts to separate the characters, either \textit{kanji}, ,\textit{hiragana} or \textit{katakana}. So we cannot learn good subwords from a small corpus by directly applying BPE or the variant \textbf{\textit{VNBPE}} on Japanese written texts. In order to learn good subwords and with a little knowledge about Japanese, we decided to use KyTea\cite{neubig2011pointwise} to do Japanese word segmentation and then apply Sennrich's BPE on those word-segmented texts. Some examples of the Japanese words going through the word segmentation and BPE are shown in Table~\ref{jbpe}. 

\begin{table}[htbp]
\begin{center}
\begin{tabular}{|l|l|l|l|}
\hline \hline
Before BPE  & After BPE & Vietnamese equi. & English equi.\\
\hline
\JP{受け入れる} & \JP{受け 入れる} & \foreignlanguage{Vietnamese}{Chấp nhận} & Accept\\
\hline
\JP{崩れ落ちる} & \JP{崩れ 落ちる} & \foreignlanguage{Vietnamese}{Thu gọn} & Collapse \\
\hline
\JP{姉ちゃん} & \JP{姉  ちゃん} & \foreignlanguage{Vietnamese}{Chị gái} & Older sister \\
\hline
\JP{哀れん} & \JP{哀  れん}  & \foreignlanguage{Vietnamese}{Đáng tiếc} & Pity \\
\hline 
\JP{取りかかる}  & \JP{取り  かかる}  & \foreignlanguage{Vietnamese}{Để bắt đầu} & To start \\
\hline \hline
\end{tabular}
\end{center}
\vspace*{-0.2cm}
\caption{\label{jbpe}Examples of Japanese words after BPE.}
\end{table}



\section{Data Augmentation}
In this section, we describe the data augmentation methods we use to increase the amount of training data in order to make our NMT systems suffer less from the low-resourced situation in Japanese $\leftrightarrow$ Vietnamese translation. Although NMT systems can predict and generate  the translation of unseen words on their vocabularies, but they only perform this well if the parallel corpus for training are sufficiently large. For many under-resourced languages, unfortunately, it hardly presents.  In reality, although the monolingual data of Vietnamese and Japanese are immensely available due to the popularity of their speakers, the bilingual Japanese-Vietnamese corpora are very limited and often in low quality or in narrowly technical domains. Therefore, data augmentation methods to exploit monolingual data for NMT systems are necessary to obtain more bilingual data, thus upgrading the translating quality.  
\subsection{Back Translation}
One of the approaches to leverage the monolingual data is to use a machine translation system  to translate those data in order to create a synthetic parallel data. Normally, the monolingual data in the target language is translated, thus the name of the method: Back Translation\cite{Sennrich2016b}. 

More specific, to generate the data for an X$\rightarrow$Y NMT system, we use the best Y$\rightarrow$X translation system we have to translate every sentence $\left\{ y_i \right\} \in Y$ in the monolingual data of language Y into sentences $\left\{x_i \right\} \in X$ in the source language X. Then we pair $\left\{ x_i , y_i \right\}$ to get the synthetic data.  Finally, original bilingual data and synthetic data are mixed to train our NMT from the start.

Back Translation can improve the estimate conditional probability  of the target word on the previous context words through adding a bilingual data with approximate translations to the source languages. Furthermore, the synthetic data might contain some translation noise from the Back Translation system, and if this noise is relevant, our NMT can be more robust in learning how to translate noisy inputs. One the other hand, if the quality of the Back Translation system is not adequate, using the synthesis data might bring adverse effects to our NMT.


In this paper,  we subsample an amount of Vietnamese monolingual data so that we can create a synthetic corpus having the same size with the Japanese-Vietnamese parallel corpus. In the end, the data we have is double in size compared to the original one.

\subsection{Mix-Source Approach}
Another data augmentation method considered useful in this low-resourced setting is the mix-source method\cite{Ha2016}. In this method, we can utilize the monolingual data of the target language in a multilingual NMT system by mixing the original source sentences with those target monolingual data. The multilingual framework then uses the share information across source and target languages to improve the decision of the target words to be chosen. 

Specifically, there are a small parallel corpus $XY$ of the language pair X-Y which has $n$ sentence pairs $\left\{ x_i , y_i \right\}$ ($i=\overline{1,n}$) and a big monolingual corpus $Y$ of the language Y which has $m$ sentences $\left\{a_j\right\}$ ($j=\overline{1,m}$, $m \gg n$). Now from the monolingual corpus $Y$ we can generate a parallel corpus $YY$ where we try to model the identical translation Y-Y: $\left\{ a_j , a_j \right\}$ ($j=\overline{1,m}$).  Then we mix $XY$ and $YY$ to get a parallel corpus of the size $m+n$. Similar to the Back Translation, we subsample $n$ Vietnamese sentence pairs from the corpus $YY$ then the size of the parallel data we have is also doubled.   

To let the NMT knows which language a certain source sentence is in and then can model the language information, we follows the conventions from \cite{Ha2016}. We append language tags to every word in both source and target sentences of the mixed $XY+YY$ corpus to indicate the language of the words. This technique shows the effectiveness in low-resourced scenarios\cite{eunah2016kit, Ha2017} and our Japanese$\leftrightarrow$Vietnamese is such a scenario. 

\section{Experiments}
\subsection{Data Preparation}
We collected Japanese-Vietnamese parallel data from TED talks extracted from WIT3's corpus\cite{cettoloEtAl:EAMT2012}. After removing blank and duplicate lines we obtained 106758 pairs of sentences. The validation set used in all experiments is {\tt dev2010} and the test set is  {\tt tst2010}.

The data augmentation methods has been applied only for the Japanese$\leftrightarrow$Vietnamese direction. For Back Translation,  we use Vietnamese monolingual data from {\tt VNESEcorpus} of DongDu\footnote{\url{http://viet.jnlp.org/download-du-lieu-tu-vung-corpus}} which includes 349578 sentences. We shuffle the lines of VNESEcorpus corpus and take out the first 106758 sentences (the same as the number of sentence pairs in the original parallel corpus).  For Mix-Source, instead of using a subsampled monolingual corpus, we use the Vietnamese part of the Japanese-Vietnamese parallel corpus in order to learn the multilingual information in the same domain.  Our datasets are listed in Table~\ref{dataset}.

\begin{table}[htbp]
\begin{center}
\begin{tabular}{|l|l|c|}
\hline \hline
\textbf{Dataset} & \textbf{Description} & \textbf{Num. of sentences} \\
\hline
Training & TED & 106758\\
\hline
Back Translation data & Subsampled DongDu & 106758 \\
Mix-Source data & Vietnamese part of TED & 106758 \\
\hline
Validation & TED dev2010 & 568  \\
Test & TED tst2010 &1220  \\
\hline \hline 
\end{tabular}
\end{center}
\caption{\label{dataset} Statistics of the dataset used in our experiments}
\vspace*{-0.5cm}
\end{table}

\subsection{Preprocessing}
After using KyTea to tokenize the Japanese texts, we learn and apply Sennrich's BPE from the tokenized texts. For  Vietnamese texts, first we use Moses scripts\footnote{\url{https://github.com/moses-smt/mosesdecoder/tree/master/scripts}} to normalize the texts from digits, punctuations and special symbols. We use {\tt pyvi}\footnote{\url{https://github.com/trungtv/pyvi}} for Vietnamese word segmentation since it is one of the best tools for this task in term of speed, robustness and performance. On the other hand, we use\textbf{\textit{VNBPE}} as an alternative way of doing word segmentation. Those two approaches are compared in an extrinsic evaluation of the NMT systems employing them. 

\subsection{System Architecture and Training}
We implement the translation systems using {\tt OpenNMT-py} framework\footnote{\url{https://github.com/OpenNMT/OpenNMT-py}} \cite{opennmt}. Our system architecture includes two bi-directional LSTM layers for the encoder and two LSTM layers for the decoder, each layer has the size of 512 hidden units. The size of source and target embedding layers is also 512. We use Adam optimizer\cite{kingma2014adam} and learning rate annealing scheme with the initial learning rate at $0.001$. We train each systems for 15 epochs with the batch size of 64. The best model in term of the unigram accuracy on the validation set is usually used to translate the test set with beam size of 16. Other settings are the default settings of {\tt OpenNMT-py}, otherwise already noted. 

\subsection{Results}
We  evaluate the quality of translation of systems based on different approaches mentioned in previous sections. The {\tt multi-BLEU} from Moses scripts is used. The results have shown in the Table~\ref{tab:results}.

\textbf{Baseline.} For the baseline systems, the training data includes KyTea-segmented Japanese texts and pyvi-segmented Vietnamese texts. For comparison purpose, we build two baseline systems for each direction: one is use the traditional phrase-based statistical machine translation (SMT), the other one is the NMT system.  Although our training set is small but we find that the NMT systems (2) are still more effective than the phrase-based SMT models (1) in both translation directions. 
   
\textbf{Subword NMT.} We applied \textbf{\textit{VNBPE}} and \textbf{\textit{JPBPE}} to the baseline's data and trained NMT systems. On Vietnamese$\leftrightarrow$Japanese, we observed an improvement of 0.6 BLEU points when we used our \textbf{\textit{VNBPE}} (3) instead of the {\tt pyvi}'s word segmentation (2). Furthermore, when we trained our NMT models using both BPE methods (4), we obtained a bigger gain of 1.15 BLEU points. The similar improvements can be found in the Japanese$\leftrightarrow$Vietnamese as well: 0.29 BLEU points between (3) and (2) and 0.57 BLEU points between (4) and (3). This draws two conclusions: (i), despite using an unsupervised Vietnamese word segmentation which is fast, robust and does not require linguistic resources, our NMT systems performed better than those systems employing a complicate word segmentation method, (ii) BPE works significantly well for Japanese texts after we tokenized the texts.

\textbf{Data Augmentation.} We use the best system for Vietnamese$\leftrightarrow$Japanese, which is the NMT systems trained on BPE-processed texts, to generate the synthetic data for Japanese$\leftrightarrow$Vietnamese translation. Although we achieved some gain (from 9.04 to 9.39 BLEU points), the effectiveness of Back Translation is not on par with its application on the translation systems of other language pairs. Looking into the Vietnamese$\leftrightarrow$Japanese translation of DongDu corpus and its BLEU score, we speculate that it is because the Vietnamese$\leftrightarrow$Japanese system is not good enough to produce reasonable synthetic data. In the meanwhile, combining Back Translation and Mix-Source brings a considerable improvements of 0.6 BLEU points compared to not using them.

\begin{table} [h]
\vspace*{-0.1cm}
\centerline{
\begin{tabular}{|c|l|c|c|}
\hline \hline
\multicolumn{4}{|c|}{ \textbf{Vietnamese$\rightarrow$Japanese}} \\ \hline
 & System  & {\tt dev2010} & {\tt tst2010}  \\
\hline
 (1) & SMT Baseline & - & 8.73 \\
 (2) & NMT Baseline & 8.68 & 9.39 \\
 (3) & + VNBPE & 9.12 & 9.89 \\
 (4) & + JPBPE & \textbf{9.74} & \textbf{11.13} \\  \hline
 \hline
\multicolumn{4}{|c|}{ \textbf{Japanese$\rightarrow$Vietnamese}} \\ \hline
& System & {\tt dev2010} & {\tt tst2010}  \\
\hline
 (1) & SMT Baseline & - & 7.73 \\
 (2) & NMT Baseline & 6.85 & 8.18 \\
 (3) & + VNBPE & 7.36 & 8.47 \\
 (4) & + JPBPE & 7.77 & 9.04 \\
 (5) & + Back Translation & 8.25 & 9.39 \\
 (6) & + Mix-Source & \textbf{8.56} & \textbf{9.64} \\  \hline
 \hline
\end{tabular}}
\caption{\label{tab:results} {Evaluation of Japanese$\leftrightarrow$Vietnamese NMT systems.}}
\vspace*{-0.4cm}
\end{table}

\section{Related Works}
Japanese-Vietnamese MT is firstly mentioned in 2005\cite{chau2005}. The authors focused on the difference from embedding structures between Japanese and Vietnamese, and then proposed rules for MT system and experiment on very small dataset (714 Japanese embedding sentences). This approach is suitable for small system applied in a specific domain or language, but it is not easily extendable to other domains or languages due to the expensiveness of building such rules. 

The other previous work for Japanese$\rightarrow$Vietnamese uses SMT\cite{Do2015}. They also conducted the experiments on parallel corpora collected from TED talks. They used phrase-based and tree-to-string models and have shown that the SMT system trained on French$\rightarrow$Vietnamese obtains better results than the system of Japanese$\rightarrow$Vietnamese because French and Vietnamese have more similarities in the structures of sentences than between Japanese and Vietnamese. We also built phrase-based systems on the TED data and achieved better BLEU scores when using NMT.

Recently, some works use monolingual data to improve the accuracy of NMT systems. \cite{Sennrich2016b} have shown significant improvements by using monolingual data on target-side to generate synthetic data and then add them to original training data.~\cite{zhang2018} have shown significant improvements by "self-learning" method to generate the target sentences based on monolingual data on the source-side and then combined them with original bilingual data to train.\cite{Ha2016} convert monolingual corpus on the target-side into a bitexts by copying target sentences to the  source sentence  and then combined  original bilingual data together on training. Our systems employ those approaches to exploit monolingual data and show the improved performance for Japanese$\leftrightarrow$Vietnamese translations.

\section{Conclusion}
We has built the first Japanese$\leftrightarrow$Vietnamese NMT systems and released the dataset as well as the associated training scripts. We have also shown that the proposed \textbf{\textit{VNBPE}} algorithm can be used for Vietnamese word segmentation in order to conduct neural machine translation. Furthermore, by adapting Back Translation and Mix-Source, our NMT systems achieved the best improvement on the dataset. In the future, we will exploit more domain and multilingual information to improve the quality of the systems.

\section{Acknowledgments}
We would like to thank the center of High-performance computing (HPC), University of Engineering and Technology, VNU, Vietnam for allowing us to use their GPUs to perform the experiments mentioned in the paper. We also thank the anonymous reviewers for their careful reading of our paper and insightful comments.

\bibliography{references}

\begin{thebibliography}{10}
\providecommand{\url}[1]{#1}
\csname url@samestyle\endcsname
\providecommand{\newblock}{\relax}
\providecommand{\bibinfo}[2]{#2}
\providecommand{\BIBentrySTDinterwordspacing}{\spaceskip=0pt\relax}
\providecommand{\BIBentryALTinterwordstretchfactor}{4}
\providecommand{\BIBentryALTinterwordspacing}{\spaceskip=\fontdimen2\font plus
\BIBentryALTinterwordstretchfactor\fontdimen3\font minus
  \fontdimen4\font\relax}
\providecommand{\BIBforeignlanguage}[2]{{%
\expandafter\ifx\csname l@#1\endcsname\relax
\typeout{** WARNING: IEEEtran.bst: No hyphenation pattern has been}%
\typeout{** loaded for the language `#1'. Using the pattern for}%
\typeout{** the default language instead.}%
\else
\language=\csname l@#1\endcsname
\fi
#2}}
\providecommand{\BIBdecl}{\relax}
\BIBdecl

\bibitem{Bahdanau2014}
\BIBentryALTinterwordspacing
D.~Bahdanau, K.~Cho, and Y.~Bengio, ``{Neural Machine Translation by Jointly
  Learning to Align and Translate},'' \emph{CoRR}, vol. abs/1409.0473, 2014.
  [Online]. Available: \url{http://arxiv.org/abs/1409.0473}
\BIBentrySTDinterwordspacing

\bibitem{Sutskever2014}
I.~Sutskever, O.~Vinyals, and Q.~V. Le, ``Sequence to sequence learning with
  neural networks,'' in \emph{Technical Report arXiv:1409.3215 [cs.CL]},
  Google. NIPS’2014, 2014.

\bibitem{wu2016google}
Y.~Wu, M.~Schuster, Z.~Chen, Q.~V. Le, M.~Norouzi, W.~Macherey, M.~Krikun,
  Y.~Cao, Q.~Gao, K.~Macherey \emph{et~al.}, ``Google's neural machine
  translation system: Bridging the gap between human and machine translation,''
  in \emph{Transactions of the Association for Computational Linguistics}, vol.
  5, pp. 339–351 2017.

\bibitem{cettolo2016iwslt}
M.~Cettolo, J.~Niehues, S.~St{\"u}ker, L.~Bentivogli, R.~Cattoni, and
  M.~Federico, ``{The IWSLT 2016 Evaluation Campaign},'' in \emph{{Proceedings
  of the 13th International Workshop on Spoken Language Translation (IWSLT
  2016)}}, Seattle, WA, USA, 2016.

\bibitem{Sennrich2016a}
R.~Sennrich, B.~Haddow, and A.~Birch, ``{Neural Machine Translation of Rare
  Words with Subword Units},'' in \emph{Association for Computational
  Linguistics (ACL 2016)}, Berlin, Germany, August 2016.

\bibitem{HochreiterLSTM}
\BIBentryALTinterwordspacing
S.~Hochreiter and J.~Schmidhuber, ``Long short-term memory,'' \emph{Neural
  Comput.}, vol.~9, no.~8, pp. 1735--1780, Nov. 1997. [Online]. Available:
  \url{http://dx.doi.org/10.1162/neco.1997.9.8.1735}
\BIBentrySTDinterwordspacing

\bibitem{Cho2014}
K.~Cho, B.~van Merrienboer, {\c{C}}.~G{\"{u}}l{\c{c}}ehre, F.~Bougares,
  H.~Schwenk, and Y.~Bengio, ``{Learning Phrase Representations using RNN
  Encoder-Decoder for Statistical Machine Translation},'' in \emph{{Proceedings
  of Eighth Workshop on Syntax, Semantics and Structure in Statistical
  Translation (SSST-8}}.\hskip 1em plus 0.5em minus 0.4em\relax Baltimore, ML,
  USA: Association for Computational Linguistics, Jule 2014.

\bibitem{gehring2017convolutional}
J.~Gehring, M.~Auli, D.~Grangier, and Y.~Dauphin, ``A convolutional encoder
  model for neural machine translation,'' in \emph{Proceedings of the 55th
  Annual Meeting of the Association for Computational Linguistics (Volume 1:
  Long Papers)}, vol.~1, 2017, pp. 123--135.

\bibitem{vaswani2017attention}
A.~Vaswani, N.~Shazeer, N.~Parmar, J.~Uszkoreit, L.~Jones, A.~N. Gomez,
  {\L}.~Kaiser, and I.~Polosukhin, ``Attention is all you need,'' in
  \emph{Advances in Neural Information Processing Systems}, 2017, pp.
  6000--6010.

\bibitem{gage1994}
P.~Gage, ``A new algorithm for data compression,'' in \emph{C Users J.,
  12(2):23–38, February}, 1994.

\bibitem{neubig2011pointwise}
G.~Neubig, Y.~Nakata, and S.~Mori, ``Pointwise prediction for robust, adaptable
  japanese morphological analysis,'' in \emph{Proceedings of the 49th Annual
  Meeting of the Association for Computational Linguistics: Human Language
  Technologies: short papers-Volume 2}.\hskip 1em plus 0.5em minus 0.4em\relax
  Association for Computational Linguistics, 2011, pp. 529--533.

\bibitem{Sennrich2016b}
R.~Sennrich, B.~Haddow, and A.~Birch, ``{Improving Neural Machine Translation
  Models with Monolingual Data},'' in \emph{Association for Computational
  Linguistics (ACL 2016)}, Berlin, Germany, August 2016.

\bibitem{Ha2016}
\BIBentryALTinterwordspacing
T.-L. Ha, J.~Niehues, and A.~Waibel, ``Toward multilingual neural machine
  translation with universal encoder and decoder,'' \emph{{Proceedings of the
  13th International Workshop on Spoken Language Translation (IWSLT 2016)}},
  2016. [Online]. Available: \url{http://arxiv.org/abs/1611.04798}
\BIBentrySTDinterwordspacing

\bibitem{eunah2016kit}
E.~Cho, J.~Niehues, T.-L. Ha, M.~Sperber, M.~Mediani, and A.~Waibel,
  ``{Adaptation and Combination of NMT Systems: The KIT Translation Systems for
  IWSLT 2016},'' in \emph{Proceedings of the 13th International Workshop on
  Spoken Language Translation (IWSLT 2016)}, Seattle, WA, USA, 2016.

\bibitem{Ha2017}
\BIBentryALTinterwordspacing
T.-L. Ha, J.~Niehues, and A.~Waibel, ``{Effective Strategies in Zero-Shot
  Neural Machine Translation},'' \emph{{Proceedings of the 14th International
  Workshop on Spoken Language Translation (IWSLT 2017)}}, 2017. [Online].
  Available: \url{http://arxiv.org/abs/1711.07893}
\BIBentrySTDinterwordspacing

\bibitem{cettoloEtAl:EAMT2012}
M.~Cettolo, C.~Girardi, and M.~Federico, ``Wit$^3$: Web inventory of
  transcribed and translated talks,'' in \emph{Proceedings of the 16$^{th}$
  Conference of the European Association for Machine Translation (EAMT)},
  Trento, Italy, May 2012, pp. 261--268.

\bibitem{opennmt}
\BIBentryALTinterwordspacing
G.~Klein, Y.~Kim, Y.~Deng, J.~Senellart, and A.~M. Rush, ``Opennmt: Open-source
  toolkit for neural machine translation,'' in \emph{Proc. ACL}, 2017.
  [Online]. Available: \url{https://doi.org/10.18653/v1/P17-4012}
\BIBentrySTDinterwordspacing

\bibitem{kingma2014adam}
D.~Kingma and J.~Ba, ``Adam: A method for stochastic optimization,''
  \emph{arXiv preprint arXiv:1412.6980}, 2014.

\bibitem{chau2005}
M.~C. Nguyen and T.~Ikeda, ``Translating japanese-vietnamese machine
  qualification structure in machine translation,'' \emph{Natural Language
  Processing}, vol.~12, no.~3, pp. 145--182, 2005.

\bibitem{Do2015}
D.~Do, M.~Utiyama, and E.~Sumita, ``Machine translation from japanese and
  french to vietnamese, the difference among language families,'' October 2015.

\bibitem{zhang2018}
Z.~Zhang, S.~Liu, M.~Li, M.~Zhou, and E.~Chen, ``Joint training for neural
  machine translation models with monolingual data,'' in \emph{Association for
  the Advancement of Artificial Copyright Intelligence (AAAI 2018)}, Louisiana,
  USA, February 2018.

\end{thebibliography}
\bibliographystyle{IEEEtran}

\end{document}